\documentclass[a4paper,10pt,twocolumn]{article}
\usepackage[a4paper,right=1.5cm,left=1.5cm,top=2cm,bottom=2.5cm,headsep=0cm,footskip=0.5cm]{geometry}
\usepackage[english,activeacute]{babel}

\usepackage[utf8]{inputenc}
\usepackage{enumitem}
\usepackage{amsmath}
\usepackage{amssymb}
\usepackage{times}
\usepackage{graphicx}
\usepackage{float}
\usepackage{csquotes} 
\usepackage{booktabs}
\usepackage{makecell}
\usepackage{tabularx}
\usepackage{multirow}
\usepackage{xcolor}
\usepackage{stfloats}
\usepackage{colortbl}
\usepackage[format=plain,labelfont=bf,textfont=it,labelsep=period]{caption}
\usepackage[colorlinks=true,allcolors=blue]{hyperref}
\usepackage{titlesec}
\usepackage{pifont}

\titleformat{\section}{\large\bfseries}{\thesection}{1em}{} %Cambia el tamaño de la fuente de las secciones a \large
\titleformat{\subsection}{\small\bfseries}{\thesubsection}{1em}{} % Cambia el tamaño de la fuente de las subsecciones a \large
\titleformat{\subsubsection}{\small\bfseries}{\thesubsubsection}{1em}{} % Cambia el tamaño de la fuente de las subsubsecciones a \large
\titlespacing{\section}{0pt}{\parskip}{\parskip}
\titlespacing{\subsection}{0pt}{\parskip}{\parskip}

\newcommand{\link}[1]{{\color{blue}\href{#1}{#1}}}
\renewcommand\thesection{\arabic{section}}
\renewcommand\thesubsection{\thesection.\arabic{subsection}}

\title{\LARGE \textbf{Zero-shot segmentation of skin tumors in whole-slide images with vision-language foundation models}}

 \author{\large Santiago Moreno* $^{1}$, Pablo Meseguer* $^{1}$, Rocío del Amor $^{1,2}$, Valery Naranjo $^{1,2}$
 \\
 \\
 \small $^1$ Instituto Universitario de Investigación en Tecnología Centrada en el Ser Humano (HUMAN-Tech), \\ \small Universitat Politécnica de Valencia, Valencia, Spain \\ \small $^2$ Artikode Intelligence S.L., Valencia, Spain \\ \small \{smorpin, pabmees, madeam2, vnaranjo\}@upv.es
 \\ \small *Contributed equally} 

\date{} %Elimina la fecha

\begin{document}	

\maketitle
\thispagestyle{empty}

\begin{center}
\large
\textbf{Abstract}
\end{center}
\small
\textit{Accurate annotation of cutaneous neoplasm biopsies represents a major challenge due to their wide morphological variability, overlapping histological patterns, and the subtle distinctions between benign and malignant lesions. Vision–language foundation models (VLMs), pre‐trained on paired image–text corpora, learn joint representations that bridge visual features and diagnostic terminology, enabling zero‐shot localization and classification of tissue regions without pixel‐level labels. However, most existing VLM applications in histopathology remain limited to slide‐level tasks or rely on coarse interactive prompts, and they struggle to produce fine‐grained segmentations across gigapixel whole‐slide images (WSIs). In this work, we introduce a Zero-shot visual-language segmentation pipeline for whole-slide images (ZEUS), a fully automated, zero‐shot segmentation framework that leverages class‐specific textual prompt ensembles and frozen VLM encoders to generate high‐resolution tumor masks in WSIs. By partitioning each WSI into overlapping patches, extracting visual embeddings, and computing cosine similarities against text prompts in order to generate a final segmentation mask. We demonstrate competitive performance on two in‐house datasets, primary spindle cell neoplasms and cutaneous metastases, highlighting the influence of prompt design, domain shifts, and institutional variability in VLMs for histopathology. ZEUS markedly reduces annotation burden while offering scalable, explainable tumor delineation for downstream diagnostic workflows.}

\normalsize

\section{Introduction}
Skin cancer ranks among the most common malignancies globally, with incidence rates continuing to raise \cite{Roky2025}. Cutaneous spindle‐cell (CSC) neoplasms form a heterogeneous group of skin tumors, including benign leiomyomas and aggressive leiomyosarcomas, which frequently confuse pathologists due to their morphological overlap \cite{Choi2018}. Cutaneous metastases (CM), often originating from breast or other organs, introduce additional diagnostic complexity, particularly when tracing the primary source of lesions \cite{Leonard2007}.

Deep learning has revolutionized computational pathology (CPath), showing promising results in tumor detection and region‐of‐interest (ROI) segmentation \cite{Danisetty2025, delAmor2022}. This accurate segmentation is crucial, as it helps identify ROIs and differentiate between tumor and healthy cells, thus improving diagnostic accuracy and downstream quantitative analysis \cite{Danisetty2025}. However, training such models is often hindered by the scarcity of labeled data and the labor-intensive process of manual pixel-level annotation, which remains a critical bottleneck in the development of robust, generalizable systems \cite{delAmor2022}. 

Foundation models, which are large-scale general-purpose AI systems trained across diverse data modalities, have recently emerged as powerful tools in CPath. These models support a wide range of downstream applications, such as rare disease classification, survival prediction, multi-stain analysis, and biomarker assessment \cite{Bilal2025}. In particular, vision-language models (VLMs) enable zero-shot classification and segmentation by linking histological imaging with associated textual data, enabling them to generalize to new tasks and reducing the reliance on exhaustive pixel-level annotations \cite{Sharma2025, Ranjbar2025}. While their performance in classification and retrieval tasks has been explored, their potential for zero-shot segmentation in complex biomedical scenarios remains under-investigated. 

Recent studies have explored zero‐shot applications of the Segment Anything Model (SAM) across diverse medical imaging modalities demonstrating its flexibility but also its dependence on user‐provided prompts for accurate region delineation \cite{Liu2024, Cui2024}. Despite these advances, accurately segmenting diagnostically relevant regions within gigapixel WSIs remains a major hurdle, especially given that CSC and CM lesions often exhibit ambiguous pathological features. In this work, we propose Zero-shot visual-language segmentation pipeline for whole-slide images (ZEUS) that automatically generates accurate segmentations without requiring any task-specific annotations, thus automating the segmentation of tumor regions in gigapixel WSIs and significantly reducing the laborious pixel-level annotation burden on pathologists. Code for using this framework is available at \link{https://github.com/cvblab/ZEUS}.

\section{Methodology}
\begin{figure*}[hbt]
  \centering
  \includegraphics[width=1.3\columnwidth]{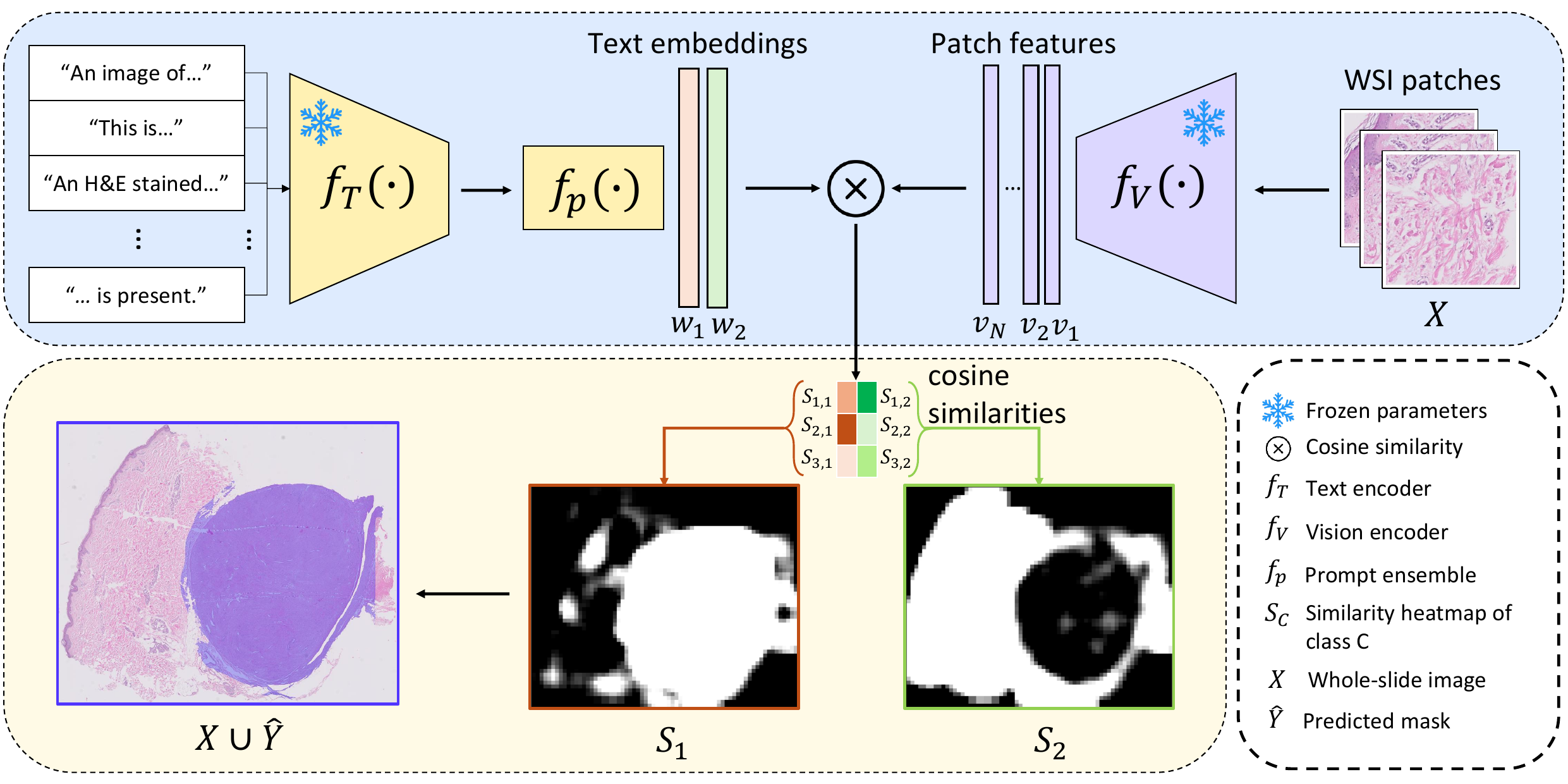}
  \caption{Overview of ZEUS workflow. \textbf{Top:} a pre-trained vision encoder (\(f_V\)) is used to extract features from WIS patches (\(v_N\)). Textual prompts for the C classes are encoded using the text encoder (\(f_T\)) to obtain the text embeddings and the ensembling function (\(f_p\)) to obtain the prompt ensembles. \textbf{Bottom:} Cosine similarities (\(s_{j,c})\) between text and patch embeddings yield class‐specific similarity maps (\(S_C\)), which are stacked and passed through a pixel‐wise \(\arg\max\) to generate the final segmentation mask \(\hat{Y}\).}
  \label{fig_ZSmil}
\end{figure*}

Figure~\ref{fig_ZSmil} illustrates our ZEUS workflow for zero‐shot ROI segmentation in gigapixel WSIs. Below, we formalize the problem and detail each pipeline component.

\textbf{Problem formulation:} WSIs are gigapixel‐scale images of tissue sections \((X\)), making direct pixel‐level processing infeasible. Our goal is to automatically segment tumor versus healthy regions in these images without any pixel‐level annotations. To this end, each WSI is first divided into \(K\) patches (tiles), such as \(X = \{x_j\}_{j=1}^K\), from which feature embeddings are extracted using a pre-trained vision encoder (\(f_V\)). Simultaneously, we define a prompt ensemble for each class. For each tile, we compute cosine similarities against all class embeddings in order to obtain the prediction masks.

\subsection{Zero-Shot Segmentation}
\textbf{Text encoding (Prompt ensembling)}: For each class \(c\), we first assemble a set of \(N\) classnames \(\{\text{classname}_n^c\}_{n=1}^N\) and a shared set of \(M\) textual templates \(\{\tau_m\}_{m=1}^M\) (e.g., “an image of \textit{CLASSNAME},” “microscopic view of \textit{CLASSNAME} cells”), each containing a placeholder for the classname. We then generate a prompt ensemble by substituting every classname into every template, producing \(N\times M\) prompts. Specifically:
\[
p_{n,m}^c = \tau_m(\text{classname}_n^c). 
\]
Each prompt is encoded through the VLM text encoder \(f_{T}(\cdot)\), and all embeddings are averaged to form a single text embedding vector:
\begin{equation}
\label{eq_wt}
w_{c} \;=\; \frac{1}{N\,M} \sum_{n=1}^{N} \sum_{m=1}^{M} f_{T}\bigl(p_{n,m}^c\bigr)\,.
\end{equation}
This prompt‐ensembling strategy captures diverse textual descriptions for each class and yields robust zero‐shot prototypes \cite{Lu2024, Zhou2024}.

\textbf{Vision encoding}: Histopathology VLMs employ a vision encoder \(f_V\) (e.g., a ViT backbone) pretrained on large collections of WSIs. For each input WSI, we first partition the tissue region into \(K\) patches. Each patch \(x_j\) is then passed through the vision encoder to obtain a \(D\)-dimensional feature vector:
\begin{equation}
\label{eq_vj}
v_j = f_V(x_j)\,
\end{equation}
These patch‐level embeddings \(\{v_j\}_{j=1}^K\) capture local morphological and textural features, forming the visual basis for downstream ROI identification and multimodal alignment.

\textbf{Multimodal alignment}: Given the set of patch embeddings and the \(C\) class ensembles obtained via Eq.~\ref{eq_wt}, we compute a similarity score for each patch–class pair via cosine similarity:
\begin{equation}
\label{eq_cosine_binary}
s_{j,c} \;=\; \frac{v_j \cdot w_{c}}{\|v_j\|\;\|w_{c}\|}, 
\quad
\end{equation}

Next, we reconstruct \(C\) per‐class similarity maps $\{S_c\}_{c=1}^C$ over the original WSI by assigning each patch’s score to all of its constituent pixels. When the patching process involves overlap, we calculate the average similarity across the overlapping pixels. Formally, if pixel $p$ belongs to tile $j$, then
\[
S_{c}(p) \;=\; s_{j,c}, 
\quad
c=1,\dots,C.
\]

Finally, we stack these \(C\) similarity maps into a single 3-dimensional array \(\displaystyle S \in \mathbb{R}^{H \times W \times C}\) and generate the segmentation mask \(\hat{Y}\) by taking a pixel‐wise argmax over the class dimension:
\begin{equation}
\label{eq_argmax_binary}
\hat{Y} = \arg\max\; (S)\
\end{equation}

\section{Experiments and results}
\subsection{Datasets}
We evaluated the ZEUS approach on two in‐house WSI collections from distinct hospitals. AI4SkIN \cite{delAmor2025} includes 90 benign leiomyoma slides from Hospital Clínico Universitario de Valencia (HCUV, 26 WSIs) and Hospital Universitario San Cecilio (HUSC, 64 WSIs). ASSIST comprises 40 secondary cutaneous metastasis WSIs, all from HCUV.

\subsection{Experimental settings}
\label{sec:exp_settings}
We evaluated two state-of-the-art histopathology VLMs —CONCH \cite{Lu2024} and KEEP \cite{Zhou2024}— for this task. Both models use a ViT‐based vision encoder and a BERT‐style text encoder, pre-trained on large image-text pairs. We first preprocessed the WSIs using CLAM \cite{Lu2021} to (1) perform tissue segmentation, (2) partition each slide into $448\times448$‐pixel patches at an effective magnification of $10\times$ with 75\% overlap, and (3) extract feature embeddings for each patch with both models. To compute cosine similarities between image features and text embeddings (Eq.~\ref{eq_cosine_binary}), we then design two separate prompt ensembles tailored to the CSC and CM identification tasks each comprising two class prototypes (“tumor” vs.\ “normal tissue”). Each ensemble consists of multiple textual templates describing the respective WSI types; we use the same number of prompts for both ensembles to ensure comparability across datasets. For evaluation of ZEUS, we report the Dice similarity coefficient (DSC) as in \cite{Silva-Rodrguez2023}, precision, and recall on both AI4SkIN and ASSIST, using pathologist-provided pixel-level annotations from HCUV and HUSC. DSC is defined as:
\begin{equation}
\mathrm{DSC} = \frac{2\,\lvert \hat{Y} \cap Y\rvert}{\lvert \hat{Y}\rvert + \lvert Y\rvert}
\end{equation}
where $\hat{Y}$ denotes the predicted binary mask—as defined in Eq.~\ref{eq_argmax_binary}—and \(Y\) the ground‐truth binary mask, with mask values of 1 indicating positive (tumor) pixels.

\subsection{Experimental results}
Table~\ref{tbl_results} summarizes the zero‐shot segmentation results. On AI4SkIN, CONCH achieves a mean DSC of $84.5\%$, while KEEP scores $82.9\%$. Although KEEP’s DSC is marginally lower, it attains higher recall at the expense of a slightly drop in precision. This suggests that KEEP has a higher tendency to over‐segment, whereas CONCH is more conservative, balancing precision and recall more evenly. In the more challenging ASSIST dataset of cutaneous metastases, the model gap widens: CONCH’s DSC falls to $57.1\%$, whereas KEEP reaches $69.5\%$. KEEP also outperforms CONCH by $5.3\%$ in precision and by $18.7\%$ in recall, with lower variability. These gains underline KEEP’s superior zero‐shot handling of heterogeneous secondary tumors.

\begin{table}[!htbp]
  \centering
  \resizebox{\columnwidth}{!}{%
    \begin{tabular}{l l c c c}
      \toprule
      \toprule
      \textbf{Dataset}
        & \textbf{Model}
        & \textbf{DSC}
        & \textbf{Precision}
        & \textbf{Recall} \\
      \midrule
      \multirow{2}{*}{\textit{AI4SkIN}}
        & CONCH \cite{Lu2024} & $0.845$\begin{tiny}$\pm0.230$\end{tiny} 
                             & $0.822$\begin{tiny}$\pm0.232$\end{tiny} 
                             & $0.922$\begin{tiny}$\pm0.200$\end{tiny} \\
        & KEEP  \cite{Zhou2024} & $0.829$\begin{tiny}$\pm0.223$\end{tiny} 
                             & $0.767$\begin{tiny}$\pm0.251$\end{tiny} 
                             & $0.961$\begin{tiny}$\pm0.160$\end{tiny} \\
      \bottomrule
      \addlinespace
      \multirow{2}{*}{\textit{ASSIST}}
        & CONCH & $0.571$\begin{tiny}$\pm0.303$\end{tiny} 
                & $0.731$\begin{tiny}$\pm0.217$\end{tiny} 
                & $0.584$\begin{tiny}$\pm0.372$\end{tiny} \\
        & KEEP  & $0.695$\begin{tiny}$\pm0.322$\end{tiny} 
                & $0.784$\begin{tiny}$\pm0.201$\end{tiny} 
                & $0.771$\begin{tiny}$\pm0.369$\end{tiny} \\
      \bottomrule
      \bottomrule
    \end{tabular}%
  }
  \caption{Zero‐shot segmentation results by model. Results denote the average and
standard deviation through all the available WSIs.}
  \label{tbl_results}
\end{table}

These foundational models are mostly pre‐trained on large cohorts of primary tumor WSIs, and their embeddings are thus most aligned to the morphological patterns of primary neoplasms \cite{Lu2024, Zhou2024}. Our results corroborate this: both CONCH and KEEP achieve high zero‐shot DSC on AI4SkIN (primary CSCs). In contrast, on ASSIST (secondary CMs) their performance drops significantly, reflecting the domain shift from primary to metastatic tissue. This differential performance highlights the need for either domain‐specific fine‐tuning or more diverse pre‐training that includes different secondary tumor morphologies to extend foundation models’ zero‐shot capabilities beyond primary cancer scenarios.

Prompt design further influences zero‐shot performance. For AI4SkIN, we used class‐specific descriptors (e.g., “benign tissue,” “leiomyoma”), which align closely with the morphology of primary CSCs and likely contribute to strong DSC performance. By contrast, ASSIST prompt ensembles employ generic classnames (e.g., “tumor,” “cancerous tissue”) without organ‐specific context, since the metastases’ primary site is unknown. This lack of anatomical precision exacerbates the semantic gap for secondary tumors and may underlie the lower DSC on ASSIST. 

In addition, in order to investigate institutional biases, we separated AI4SkIN segmentation metrics by source hospital (Table~\ref{tbl_hospitals}). Slides from HUSC consistently outperform those from HCUV by almost $8\%$ on DSC, ratifying prior reports of center‐specific domain shifts in digital pathology models \cite{deJong2025}. Differences in staining protocols or scanner characteristics can thus materially affect zero‐shot segmentation performance. Moreover, inherent variability in pathologist annotations further compounds these discrepancies, as inter‐observer differences introduce additional noise into the ground‐truth masks. This center‐dependent variability highlights the need for center‐agnostic pre‐training or targeted domain adaptation to ensure reliable deployment of foundation models across diverse clinical settings.

\begin{table}[!htbp]
  \centering
  \resizebox{\columnwidth}{!}{%
    \begin{tabular}{l l c c c}
      \toprule
      \toprule
      \textbf{Dataset}
        & \textbf{Hospital}
        & \textbf{DSC}
        & \textbf{Precision}
        & \textbf{Recall} \\
      \midrule
      \multirow{2}{*}{\textit{AI4SkIN}}
        & HCUV & $0.780$\begin{tiny}$\pm0.167$\end{tiny} 
               & $0.683$\begin{tiny}$\pm0.197$\end{tiny} 
               & $0.968$\begin{tiny}$\pm0.082$\end{tiny} \\
        & HUSC & $0.860$\begin{tiny}$\pm0.241$\end{tiny} 
               & $0.839$\begin{tiny}$\pm0.244$\end{tiny} 
               & $0.931$\begin{tiny}$\pm0.208$\end{tiny} \\
      \bottomrule
      \bottomrule
    \end{tabular}%
  }
  \caption{Center-specific zero-shot segmentation performance on AI4SkIN. Results denote the mean and standard deviation across all WSIs within each center.}%, considering metrics from both models.}
  \label{tbl_hospitals}
\end{table}

\subsection{Qualitative analysis} 
\begin{figure}[!htbp]
  \centering
  \includegraphics[width=\columnwidth]{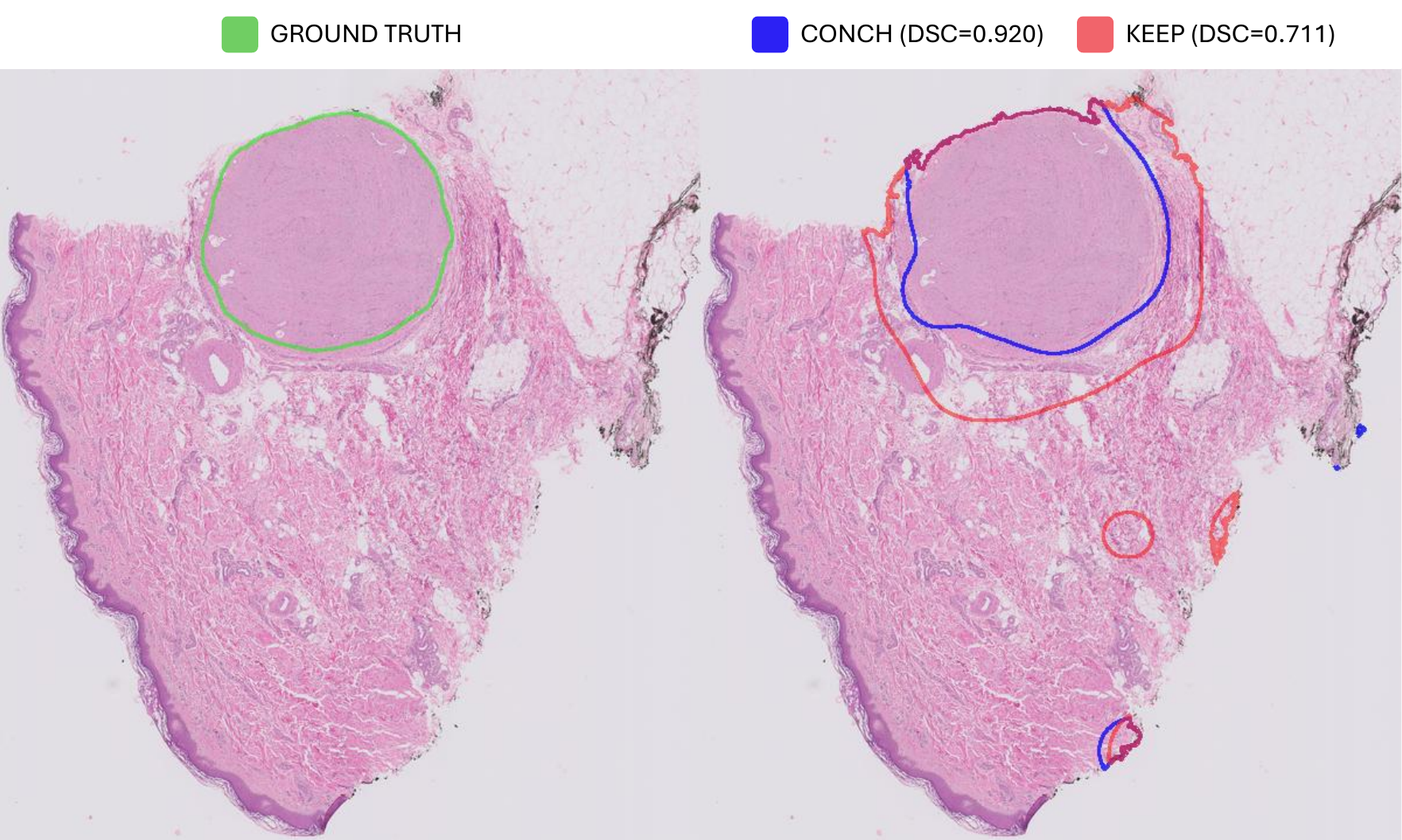}
  \caption{Contours of the predicted segmentation masks for both models CONCH (blue) and KEEP (red) with their respective DSC value, and pathologist annotation (green) on a HUSC leiomyoma WSI.}
  \label{fig_qualitative}
\end{figure}

Recent zero‐shot segmentation methods often rely on interactive guidance—such as user‐selected points or bounding boxes—to delineate tumor regions in histopathology images \cite{Liu2024, Cui2024}. While effective, these approaches still demand manual input and cannot be fully automated. By contrast, our framework uses only textual prompts to drive zero‐shot segmentation without any further human intervention. In Figure~\ref{fig_qualitative}, we show a representative AI4SkIN WSI crop overlaid with automatically generated masks from both models alongside the ground‐truth annotation. The segmentation closely follows the pathologist’s contour of the leiomyoma, illustrating that zero-shot configurations for VLMs can localize RoIs. %even in gigapixel WSIs. 

Despite these competitive results, our method inherits the limitations of patch-level processing. Since each $448\times448$ patch is far larger than a single pixel, the resulting lesion contours exhibit block-like artifacts and may not capture fine boundary details, which is reflected in the results obtained from smaller dimension tumors and those with irregular shapes. To mitigate this, we applied a 75\% patch overlap during WSI partitioning as described in Section~\ref{sec:exp_settings}, which effectively creates 'virtual' sub-patches along patch edges and results in smoother and more accurate contours without changing the overall patch size.

\section{Conclusion}
In this paper, we explored zero‐shot segmentation in digital pathology by leveraging VLMs capabilities. We have demonstrated that carefully curated textual prompt sets can be effective semantic anchors for automated identification of complex ROIs. ZEUS pipeline runs completely at inference time, partitioning gigapixel whole‐slide images into manageable patches, extracting high‐level features, and matching against class ensembles via cosine similarity. Through quantitative and qualitative analysis, we observed that performance varies according to dataset characteristics, primary tumors benefit from precise tissue‐specific prompts, while heterogeneous metastases challenge straightforward zero‐shot approaches. We also observed how institutional biases, including variability in staining and scanner differences, affect consistency in segmentation between medical centers. Looking ahead, integrating multi‐scale context, refining boundary representations, and incorporating domain‐adaptive paradigms will be key steps toward robust and generalizable segmentation tools.

% \small
% \textbf{Funding:} This work has received funding from the Spanish Ministry of Economy and Competitiveness through the project PID2022-140189OB-C21 (ASSIST) and from the Generalitat Valenciana through project COM-TACTS2 (CIPROM/2022/20).

% ---- Bibliography ----
\begin{small}
\bibliographystyle{ieeetr}
\bibliography{references}
\end{small}

\end{document}